\documentclass[10pt,twocolumn,letterpaper]{article}

\usepackage{cvpr}
\usepackage{times}
\usepackage{epsfig}
\usepackage{graphicx}
\usepackage{amsmath}
\usepackage{amssymb}
\usepackage{array}

\usepackage{mmstyle}
\usepackage{booktabs} 
\usepackage{algorithm}
\usepackage{algorithmic}
\usepackage{diagbox}
\usepackage{multirow}

\makeatletter
\def\moverlay{\mathpalette\mov@rlay}
\def\mov@rlay#1#2{\leavevmode\vtop{%
		\baselineskip\z@skip \lineskiplimit-\maxdimen
		\ialign{\hfil$\m@th#1##$\hfil\cr#2\crcr}}}
\newcommand{\charfusion}[3][\mathord]{
	#1{\ifx#1\mathop\vphantom{#2}\fi
		\mathpalette\mov@rlay{#2\cr#3}
	}
	\ifx#1\mathop\expandafter\displaylimits\fi}
\makeatother
\newcommand{\bigcupdot}{\charfusion[\mathop]{\bigcup}{+}}

\usepackage[dvipsnames]{xcolor}

\usepackage[pagebackref=true,breaklinks=true,letterpaper=true,colorlinks,bookmarks=false]{hyperref}

\cvprfinalcopy 

\def\cvprPaperID{524} 

\ifcvprfinal\fi
\begin{document}
	
\title{A Local-to-Global Approach to Multi-modal Movie Scene Segmentation}
		
	\author{Anyi Rao$^{1}$, Linning Xu$^{2}$, Yu Xiong$^{1}$, Guodong Xu$^{1}$, Qingqiu Huang$^{1}$, Bolei Zhou$^{1}$, Dahua Lin$^{1}$\\
		$^{1}$CUHK - SenseTime Joint Lab, The Chinese University of Hong Kong\\
		$^{2}$The Chinese University of Hong Kong, Shenzhen \\
		{\tt\small \{anyirao, xy017, xg018, hq016, bzhou, dhlin\}@ie.cuhk.edu.hk, linningxu@link.cuhk.edu.cn}
	}

\maketitle
\label{key}


\begin{abstract}

Scene, as the crucial unit of storytelling in movies, contains complex activities 
of actors and their interactions in a physical environment. 
Identifying the composition of scenes serves as a critical step towards 
semantic understanding of movies. 
This is very challenging -- compared to the videos studied in conventional vision 
problems, \eg~action recognition, as scenes in movies usually contain much richer
temporal structures and more complex semantic information. 
Towards this goal, we scale up the scene segmentation task by 
building a large-scale video dataset \textit{MovieScenes},  
which contains $21K$ annotated scene segments from $150$ movies. 
We further propose a local-to-global scene segmentation framework,
which integrates multi-modal information across three levels, 
\ie~clip, segment, and movie. 
This framework is able to distill complex semantics from hierarchical temporal structures over a long movie, providing 
top-down guidance for scene segmentation.
Our experiments show that the proposed network is able to segment a movie 
into scenes with high accuracy, consistently outperforming previous methods.
We also found that pretraining on our MovieScenes can 
bring significant improvements to the existing approaches. 
\footnote{The dataset will be published 
in compliance with regulations. 
Intermediate features, pretrained models and related codes will be released. \url{https://anyirao.com/projects/SceneSeg.html}}
\end{abstract}


\section{Introduction}
\label{sec:introduction}

\begin{figure}[!t]
	\begin{center}
		\includegraphics[width=\columnwidth]{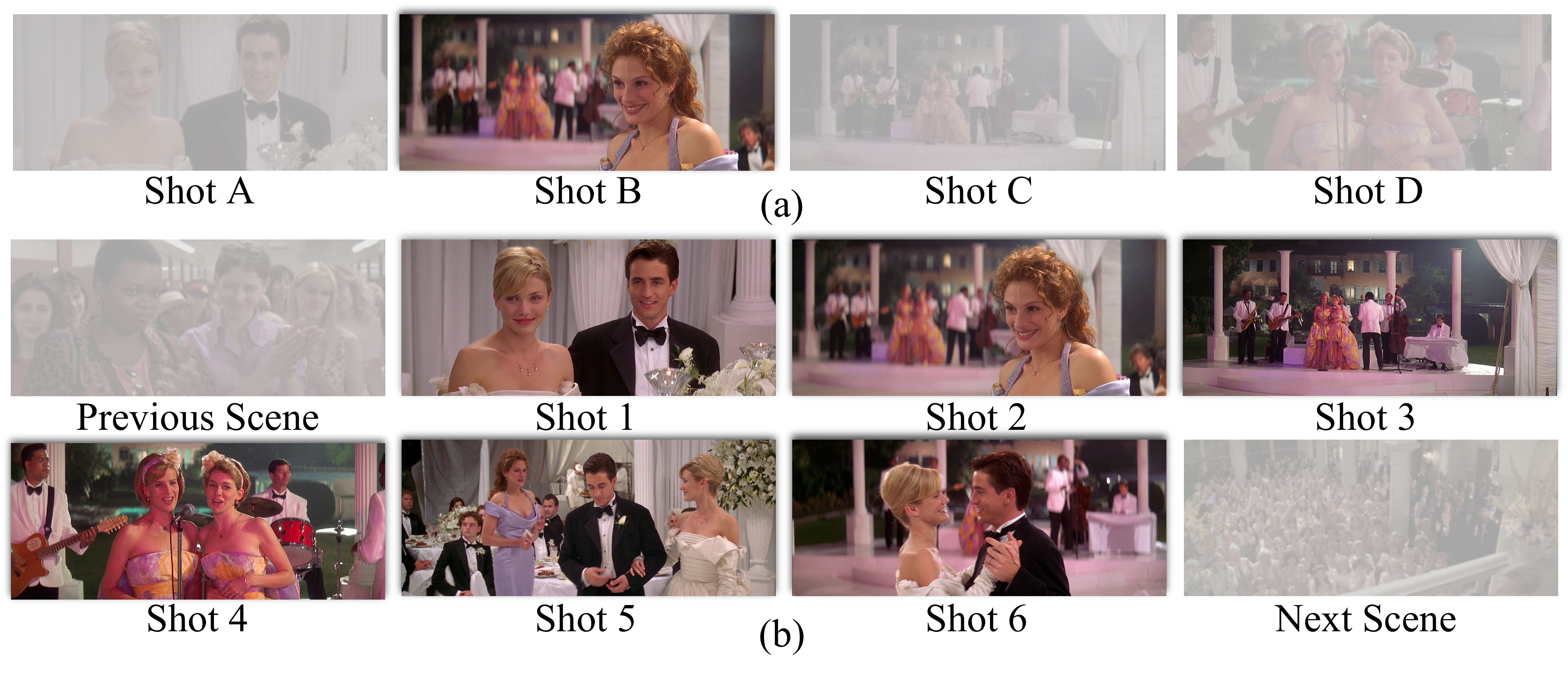}
	\end{center}
	\vspace{-15pt}
	\caption{\small
		When we look at any single shot from figure (a), \eg the woman in shot B, we cannot infer what the current event is. Only when we consider all the shots 1-6 in this scene, as shown in figure (b), we can recognize that ``this woman is inviting a couple to dance with the band.''
	}
	\label{fig:teaser}
	\vspace{-10pt}
\end{figure}

%
Imagine you are watching the movie \textit{Mission Impossible} starred 
by Tom Cruise: In a fight scene, Ethan leaps onto a helicopter's landing 
skid and attaches an exploding gum to the windshield to destroy the enemy. 
Suddenly, the story jumps into an emotional scene where Ethan pulled the trigger 
and sacrificed his life to save his wife Julia. 
Such a dramatic change of scenes plays an important role in the movie's storytelling. 
Generally speaking, a movie is composed of a well-designed series of intriguing scenes 
with transitions, where the underlying storyline determines the order of the scenes 
being presented.
Therefore recognizing the movie scenes, including the detection of scene boundaries 
and the understanding of the scene content, facilitates a wide-range of 
movie understanding tasks such as scene classification, 
cross movie scene retrieval, human interaction graph and 
human-centric storyline construction. 

%
It is worth noting that \emph{scenes} and \emph{shots} are essentially different. 
In general, a \emph{shot} is captured by a camera that operates for 
an uninterrupted period of time and thus is \emph{visually continuous}; 
while a \emph{scene} is a semantic unit at a higher level.
As illustrated in Figure~\ref{fig:teaser}, a scene comprises a sequence of shots 
to present a \emph{semantically coherent} part of the story. 
Therefore, whereas a movie can be readily divided into shots based on 
simple visual cues using existing tools~\cite{sidiropoulos2011temporal}, the task of identifying 
those sub-sequences of shots that constitute scenes is challenging,
as it requires semantic understanding in order to discover
the associations between those shots that are semantically consistent but 
visually dissimilar. 

%
There has been extensive studies on video understanding.
Despite the great progress in this area, most existing works focus on recognizing
the categories of certain activities from short videos~\cite{wang2016temporal,caba2015activitynet,monfort2019moments}. 
More importantly, these works assume a list of pre-defined categories that are 
visually distinguishable. 
However, for movie scene segmentation, it is impossible to have such a list of 
categories. Additionally, shots are grouped into scenes according to their
semantical coherence rather than just visual cues. 
Hence, a new method needs to be developed for this purpose.

To associate visually dissimilar shots, we need
semantical understanding. The key question here is 
{\em ``how can we learn semantics without category label?''}
Our idea to tackle this problem consists in three aspects:
1) Instead of attempting to categorize the content, we focus on scene \emph{boundaries}. 
We can learn what constitute a boundary between scenes in a supervised way, 
and thus get the capability of differentiating between within-scene and cross-scene 
transitions.
2) We leverage the cues contained in multiple semantic elements, including \emph{place}, \emph{cast}, 
\emph{action}, and \emph{audio}, to identify the associations across shots. 
By integrating these aspects, we can move beyond visual observations and establish 
the semantic connections more effectively. 
3) We also explore the top-down guidance from the overall understanding of 
the movie, which brings further performance gains.

%
Based on these ideas, we develop a local-to-global framework that performs
scene segmentation through three stages:
1) extracting shot representations from multiple aspects, 
2) making local predictions based on the integrated information, 
and finally 
3) optimizing the grouping of shots by solving a global optimization problem.
To facilitate this research, we construct \emph{MovieScenes},
a large-scale dataset that contains over $21K$ scenes containing
over $270K$ shots from $150$ movies. 

%
Experiments show that our method raise performance by $68\%$ (from $28.1$ to $47.1$ in terms of average precision) than the existing best method~\cite{baraldi2015deep}.
Existing methods pretrained on our dataset also have a large gain in performance.


\section{Related Work}
\label{sec:related}

\noindent\textbf{Scene boundary detection and segmentation.}
The earliest works exploit a variety of unsupervised methods.
\cite{rui1998exploring}~clusters shots according to shot color similarity. 
In~\cite{rasheed2003scene}, the author plots a shot response curve from low-level visual features and set a threshold to cut scene.
\cite{chasanis2008scene,brandon2018}~further group shots using spectral clustering with a fast global k-means algorithm.
\cite{han2011video,tapaswi2014storygraphs} predict scene boundaries with dynamic programming by optimizing a predefined optimizing objective.
Researchers also resort to other modality information, \eg~\cite{liang2009novel}~leverages scripts with HMM, \cite{sidiropoulos2011temporal}~uses low-level visual and audio features to build scene transition graph.
These unsupervised methods are not flexible and heavily rely on 
manually setting parameters for different videos.

Researchers move on to supervised approaches and start to build up new datasets. \emph{IBM OVSD}~\cite{rotman2017optimal} consists of 21 short videos with rough scenes, which may contain more than one plot.
\emph{BBC Planet Earth}~\cite{baraldi2015deep} comes from 11 Episodes of BBC documentaries. \cite{protasov2018using}~generates synthetic data from \emph{Places205}~\cite{zhou2018places}. 
However, the videos in these datasets lack rich plots or storylines, thus limits their real-world applications. The number of test videos is so small that cannot reflect the effectiveness of the methods considering the vast variety of scenes.
Additionally, their methods take shot as the analytical unit and implement scene segmentation in the local region recursively. Due to their lack of consideration of the semantics within a scene, it is hard to learn high-level semantics and achieve an ideal result.

\vspace{-10pt}
\paragraph{Scene understanding in images and short videos.}
Image-based scene analysis
~\cite{zhou2018places,yatskar2016,gupta2015visual}
can infer some basic knowledge about scenes, \eg~what is contained in this image.
However, it is hard to tell the action from a single static image since it lacks contextual information around it.
Dynamic scene understanding are further studied with seconds-long short videos~\cite{caba2015activitynet,monfort2019moments}. 
However, all these videos take single shot video without enough variations capturing the change of time and places compared to long videos.

\vspace{-10pt}
\paragraph{Scene understanding in long videos.}
There are few datasets focusing on scene in long videos.
Most available long video datasets focus on identifying casts in movies or TV series~\cite{bojanowski2013finding,huang2018unifying,ramanathan2014linking} and localizing and classifying the actions~\cite{gu2018ava}.
\emph{MovieGraphs}~\cite{vicol2018moviegraphs} focuses on the individual scene clips in a movie and the language structures of a scene. 
Some transition parts between scenes are discarded, making the information incomplete.

In order to achieve more general scene analysis that could be extended to videos with long time duration, we address scene segmentation in movies with our large-scale \textit{MovieScenes} dataset. 
We propose a framework considering both the relationship among shots locally and the relationship among scenes globally using multiple semantic elements, achieving much better segmentation results. 
\section{MovieScenes Dataset}
\label{sec:dataset}

To facilitate the scene understanding in movies, we construct
\emph{MovieScenes},
a large-scale scene segmentation dataset that contains $21K$ scenes derived by
grouping over $270K$ shots from $150$ movies.
This dataset provides a foundation for studying the complex semantics within scenes, and facilitates plot-based long video understanding on the top of scenes. 

\begin{figure*}[!t]
	\begin{center}
		\includegraphics[width=\linewidth]{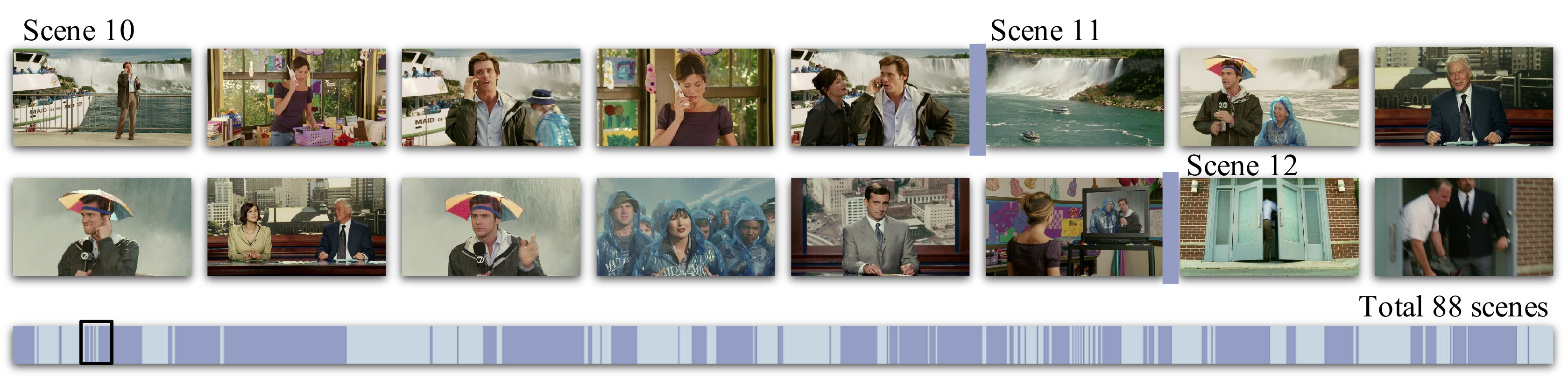}
	\end{center}
	\vspace{-5pt}
	\caption{\small
		Example of the annotated scenes from movie \textit{Bruce Almight (2003)}. The blue line in the bottom corresponds to the whole movie timeline where the dark blue and light blue regions represent different scenes. In Scene 10, the characters are having a phone call in two different places, thus it requires a semantic understanding of this scene to prevent it from categorizing them into different scenes. In Scene 11, the task becomes even more difficult, as this live broadcasting scene involves more than three places and groups of characters. In this case, visual cues only are likely to fail, thus the inclusion of other aspects such as the audio cues becomes critical.
	}
	\label{fig:montage}
	\vspace{-4pt}
\end{figure*}

\subsection{Definition of Scenes}
Following previous definition of \emph{scene}~\cite{rasheed2003scene,chasanis2008scene,han2011video,tapaswi2014storygraphs}, 
a scene is a plot-based semantic unit, where a certain activity takes place among a certain group of characters. 
While a scene often happens in a fixed place, it is also possible that a scene traverses between multiple places continually, \eg~during a fighting scene in a movie, the characters move from indoor to outdoor. These complex entanglements in scenes cast more difficulty in the accurate detection of scenes which require high-level semantic information.
Figure~\ref{fig:montage} illustrates some examples of annotated scenes in \textit{MovieScenes}, demonstrating this difficulty.

The vast diversity of movie scenes makes it hard for the annotators complying with each other.
To ensure the consistency of results from different annotations,
during the annotation procedure, we provided a list of ambiguous examples with specific guidance to clarify how such cases should be handled. Moreover, all data are annotated by different annotators independently for multiple times. In the end, our multiple times annotation with the provided guidance leads to highly consistent results,~\ie $89.5\%$ high consistency cases in total, as shown in Table~\ref{tab:dataanno}.

\begin{table}[!t]
	\caption{
		Data consistency statistics of \emph{MovieScenes}. We divide all annotations into three categories: \emph{high/low} {consistency cases} and \emph{Unsure} {cases} according to annotators consistency. \emph{unsure} {cases} are discarded in our experiments. More details are specified in the supplementary materials.}
	\vspace{2pt}
	\resizebox{\columnwidth}{!}{%
		\begin{tabular}{lrrrr}
			\toprule
			Consist. &  High & Low  & Unsure \\ \midrule
			Transit.  & \ \ 16,392  (76.5\%) & \ \ \ \ 5,036 (23.5\%) & -           \\
			Non-trans.    & 225,836 (92.6\%) 	   & 18,048 (7.4\%)       & -           \\
			Total       & 242,052 (89.5\%)     & 23,260 (8.6\%)      & 5,138 (1.9\%)        \\
			\bottomrule
		\end{tabular}
	}
	\label{tab:dataanno}
	\vspace{-0pt}
\end{table}

\begin{table}[!t]
	\caption{
		A comparison of existing scene datasets.}
	\vspace{-8pt}
	\begin{center}
		\resizebox{\columnwidth}{!}{
			\begin{tabular}{lrrrrl}
				\toprule
				& \#Shot     & \#Scene & \#Video & Time(h) & Source         \\ \midrule
				OVSD~\cite{rotman2017optimal}            & 10,000 & 300       & 21        & \ \ 10                        & MiniFilm         \\
				BBC~\cite{baraldi2015deep}                  &4,900        & 670       & 11        &\ \ \ \ 9                         & Docu.         \\
				\textit{MovieScenes}                & 270,450      & 21,428    & 150       & 297                      & Movies        \\
				\bottomrule
			\end{tabular}
		}
	\end{center}
	\label{tab:datacomp}
	\vspace{-15pt}
\end{table}

\subsection{Annotation Tool and Procedure}
Our dataset contains $150$ movies, and it would be a prohibitive amount of work if the annotators go through the movies frame by frame.
We adopt an shot-based approach, based on the understanding that a \emph{shot}\footnote{A shot is an unbroken sequence of frames recorded from the same camera.} could always be uniquely categorized into one scene. Consequently, the scene boundaries must be a subset of all the shot boundaries. For each movie, we first divide it into shots using off-the-shelf methods~\cite{sidiropoulos2011temporal}. This shot-based approach greatly simplifies the scene segmentation task and speeds up the annotation process. We also developed a web-based annotation tool\footnote{A demonstrated figure of UI is shown in supplementary materials.} to facilitate annotation. All of the annotators went through two rounds annotation procedure to ensure the high consistency. In the first round, we dispatch each chunk of movies to three independent annotators for later consistency check. In the second round, inconsistent annotations will be re-assigned to two additional annotators for extra evaluations.

\subsection{Annotation Statistics}
\noindent\textbf{Large-scale.}
Table~\ref{tab:datacomp} compares \emph{MovieScenes} with existing similar 
video scene datasets. We show that \emph{MovieScenes} is significantly larger than other datasets in terms of the number of shots/scenes and the total time duration.
Furthermore, our dataset covers a much wider range of diverse sources of data, capturing all kinds of scenes, compared with short films or documentaries.

\vspace{4pt}
\noindent\textbf{Diversity.}
Most movies in our dataset have time duration between $90$ to $120$ minutes, providing rich information about individual movie stories. 
A wide range of genres is covered, including most popular ones such as dramas, thrillers, action movies, making our dataset more comprehensive and general.
The length of the annotated scenes varies from less than $10s$ to more than $120s$, where the majority last for $10\sim30s$. This large variability existing in both the movie level and the scene level makes movie scene segmentation task more challenging.\footnote{More statistical results are specified in the supplements.}

\section{Local-to-Global Scene Segmentation}
As mentioned above, a scene is a series of continuous shots.
Therefore, scene segmentation can be formulated as a binary classification problem,
\ie to determine whether a shot boundary is a scene boundary.
However, this task is not easy, since segmenting scenes requires the recognition of multiple semantic aspects and usage of the complex temporal information.

To tackle this problem, 
we propose a Local-to-Global Scene Segmentation framework (LGSS). The overall formulation is shown in Equation~\ref{eq:formulation}.
A movie with $n$ shots is represented as a shot sequence $[\mathbf{s}_1, \cdots,\mathbf{s}_n]$, where each shot is represented with multiple semantic aspects.
We design a three-level model to incorporate different levels of contextual information, \ie clip level ($\mathcal{B}$), segment level ($\mathcal{T}$) and movie level ($\mathcal{G}$), based on the shot representation $\mathbf{s}_i$. Our model gives a sequence of predictions $[o_1, \cdots, o_{n-1}]$, where $o_i \in \{0, 1\}$ denotes whether the boundary between the $i$-th and ($i+1$)-th shots is a scene boundary.
\begin{equation}
	\begin{aligned}
		\mathcal{G}\{\mathcal{T}[\mathcal{B}([\mathbf{s}_1,\mathbf{s}_2,\cdots,\mathbf{s}_{n}])]\} = [o_1,o_2,\cdots,o_{n-1}]
	\end{aligned}
	\label{eq:formulation}
\end{equation}

\begin{figure*}[!t]
	\begin{center}
		\includegraphics[width=\linewidth]{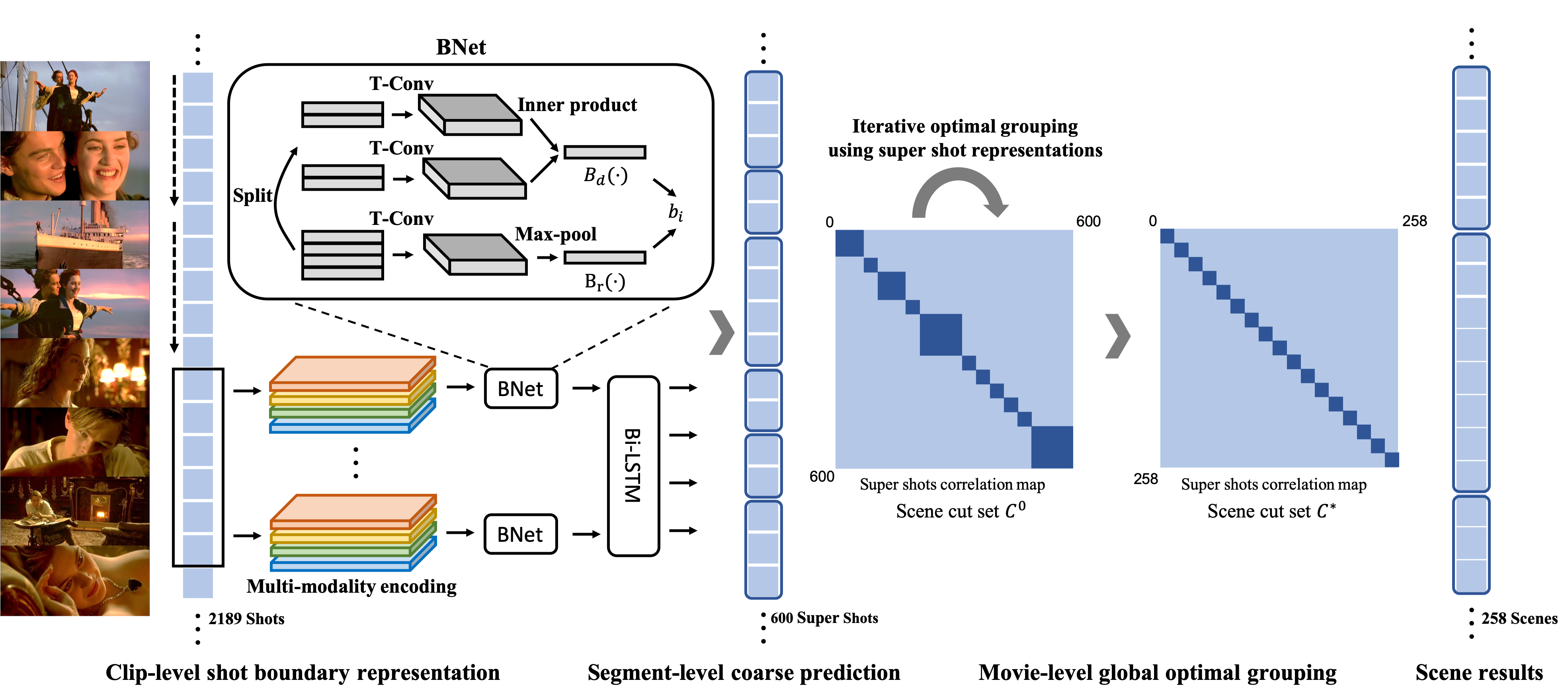}
	\end{center}
	\caption{\small
		Local-to-Global Scene Segmentation framework (LGSS). At the clip level, we extract four encoding for each shot and take a BNet to model shot boundary. The local sequence model outputs a rough scene cut results at the segment level. Finally, at the movie level, global optimal grouping is applied to refine the scene segmentation results.
	}
	\label{fig:model}
	\vspace{8pt}
\end{figure*}

In the following parts of this section, we will first introduce how to get $\mathbf{s}_i$, namely how to represent the shot with multiple semantic elements.
Then we will illustrate the details of the three levels of our model, \ie $\mathcal{B}$, $\mathcal{T}$ and $\mathcal{G}$.
The overall framework is shown in Figure~\ref{fig:model}.

\subsection{Shot Representation with Semantic Elements} 
Movie is a typical multi-modal data that contains different high-level semantic elements.
A global feature extracted from a shot by a neural network, which is widely used by previous works~\cite{baraldi2015deep,tapaswi2014storygraphs}, is not enough to capture the complex semantic information.

A scene is where a sequence of shots sharing some common elements, \eg place, cast, \etc.
Thus, it is important to take these related semantic elements into consideration for better shot representation.
In our LGSS framework, a shot is represented with four elements that play important roles in the constitution of a scene, namely \textit{place}, \textit{cast}, \textit{action}, and \textit{audio}.

To obtain semantic features for each shot $\mathbf{s}_i$, we utilize 1) ResNet50~\cite{he2016deep} pretrained on Places dataset~\cite{zhou2018places} on key frame images to get \textit{place} features, 2) Faster-RCNN~\cite{NIPS2015_5638} pretrained on CIM dataset~\cite{huang2018unifying} to detect cast instances and ResNet50 pretrained on PIPA dataset~\cite{zhang2015beyond} to extract \textit{cast} features, 3) TSN~\cite{TSN2016ECCV} pretrained on AVA dataset~\cite{gu2018ava} to get \textit{action} features, 4) NaverNet~\cite{chung2019naver} pretrained on AVA-ActiveSpeaker dataset~\cite{roth2019ava} to separate speech and background sound, and stft~\cite{umesh1999fitting} to get their features respectively in a shot with 16K Hz sampling rate and 512 windowed signal length, and concatenate them to obtain \textit{audio} features.

\subsection{Shot Boundary Representation at Clip Level}
As we mentioned before, scene segmentation can be formulated as a binary classification problem on shot boundaries.
Therefore, how to represent a shot boundary becomes a crucial question.
Here, we propose a Boundary Network (BNet) to model the shot boundary.
As shown in Equation~\ref{eq:bnet}, BNet, denoted as $\mathcal{B}$,
takes a clip of the movie with $2w_b$ shots as input and outputs a boundary representation $\vb_i$.
Motivated by the intuition that a boundary representation should capture both the \emph{differences} and the \emph{relations} between the shots before and after,
BNet consists of two branches, namely $\mathcal{B}_d$ and $\mathcal{B}_r$.
$\mathcal{B}_d$ is modeled by two temporal convolution layers, each of them embeds the shots before and after the boundary respectively, following an inner product operation to calculate their differences.
$\mathcal{B}_r$ aims to capture the relations of the shots, it is implemented by a temporal convolution layer followed a max pooling.
\begin{equation}
	\begin{aligned}
		\vb_i & = \mathcal{B}([\mathbf{s}_{i-(w_b-1)},\cdots,\mathbf{s}_{i+w_b}]) \quad (\text{window size } 2 w_b)\\
		& = 
		\left[
		\begin{array}{l}
		\mathcal{B}_d([\mathbf{s}_{i-(w_b-1)},\cdots,\mathbf{s}_i], [\mathbf{s}_{i+1},\cdots,\mathbf{s}_{i+w_b}]) \\
		\mathcal{B}_r([\mathbf{s}_{i-(w_b-1)},\cdots, \mathbf{s}_i \; , \; \mathbf{s}_{i+1}, \cdots, \mathbf{s}_{i+w_b}])
		\end{array}
		\right] 
	\end{aligned}
	\label{eq:bnet}
\end{equation}

\subsection{Coarse Prediction at Segment Level}
After we get the representatives of each shot boundary $\vb_i$, the problem becomes predicting a sequence binary labels $[o_1,o_2,\cdots,o_{n-1}]$ based on the sequence of representatives $[\vb_1, \cdots, \vb_{n-1} ]$,
which can be solved by a sequence-to-sequence model~\cite{graves2005framewise}.
However, the number of shots $n$ is usually larger than $1000$, which is hard for existing sequential models to contain such a long memory.
Therefore, we design a segment-level model to predict a coarse results based on a movie segment that consists of $w_t$ shots ($w_t \ll n$).
Specifically, we use a sequential model $\mathcal{T}$, \eg a Bi-LSTM~\cite{graves2005framewise}, with stride $w_t / 2$ shots to predict a sequence of coarse score  $[p_1, \cdots, p_{n-1} ]$, as shown in Equation~\ref{eq:seq-model}.
Here $p_i \in [0,1]$ is the probability of a shot boundary to be a scene boundary.
\begin{equation}
[p_1, \cdots, p_{n-1} ] = \mathcal{T}([\vb_1, \cdots, \vb_{n-1}])
\label{eq:seq-model}
\end{equation}

Then we get a coarse prediction $\bar{o}_i \in \{0, 1\}$,
which indicates whether the $i$-th shot boundary is a scene boundary.
By binarizing $p_i$ with a threshold $\tau$, we get
\begin{equation}
\bar{o}_i =
	\left\{
	\begin{array}{l}
	1 \qquad \text{if~~} p_i > \tau, \\
	0 \qquad \text{otherwise}.
	\end{array}
	\right.
	\label{eq:bin}
\end{equation}

\subsection{Global Optimal Grouping at Movie Level}
The segmentation result $\bar{o}_i$ obtained by the segment-level model $\mathcal{T}$ is not good enough, since it only considers the local information over $w_t$ shots while ignoring the global contextual information over the whole movie.
In order to capture the global structure, we develop a global optimal model $\mathcal{G}$ to take movie-level context into consideration.
It takes the shot representations $\mathbf{s}_i$ and the coarse prediction $\bar{o}_i$ as inputs and make the final decision $o_i$ as follows,
\begin{equation}
[o_1, \cdots, o_{n-1} ] = \mathcal{G}([\mathbf{s}_1, \cdots, \mathbf{s}_{n}], [\bar{o}_1, \cdots, \bar{o}_{n-1}])
\label{eq:gloal-opt}
\end{equation}

The global optimal model $\mathcal{G}$ is formulated as an optimization problem.
Before introducing it, we establish the concept of super shots and objective function first.  

The local segmentation gives us an initial rough scene cut set $\mC =\{\cC_k\}$, here we denote $\cC_k$ as a \emph{super shot}, \ie a sequence of consecutive shots determined by the segment-level results $[\bar{o}_1, \cdots, \bar{o}_{n-1}]$.
Our goal is to merge these super shots 
into $j$ scenes $\mPhi(n=j) = \{\phi_1, \dots, \phi_j\}$, where $\mC = \bigcupdot_{k=1}^j \phi_k$ and $|\phi_k| \geq 1$.
Since $j$ is not given, to automatically decide the target scene number $j$,
we need to look through all the possible scene cuts, 
\ie~$F = \max_{j, j<|\mC|} F(n=j)$.
With fixed $j$, we want to find the optimal scene cut set $\mPhi^\star(n=j)$. 
The overall optimization problem is as follows,
\begin{align} \label{eq:optimf}
F^{\star} &= 
\max_{j} F(n=j) \\
& \nonumber = \max_{j} \left(\max_{\mPhi} \sum_{\phi_k \in \mPhi}   g(\phi_k) \right),  \\
& \nonumber \textrm{s.t.} \quad j<|\mC|, |\mPhi| = j.
\end{align}
Here, $g(\phi_k)$ is the optimal scene cut score achieved by the scene $\phi_k$. It formulates the relationship between a super shot $\cC_l \in \phi_k$ and the rest super shots $\mathcal{P}_{k,l} = \phi_{k}\backslash \cC_l$. 
$g(\phi_k)$ constitutes two terms to capture a global relationship and a local relationship, $F_s(\cC_k,\mathcal{P}_{k})$ is similarity score between $\cC_k$ and $\mathcal{P}_{k}$,
and $F_t(\cC_k,\mathcal{P}_{k})$ is an indicate function that whether there is a very high similarity between $\cC_k$ and any super shot from $\mathcal{P}_{k}$ aiming to formulate shots thread in a scene. Specifically,
\vspace{2pt}
\begin{align}
 \nonumber g(\phi_{k}) = 
\sum_{\cC_k \in \phi_k} 
f(\cC_k,\mathcal{P}_{k}) = \sum_{\cC_k \in \phi_k}  (F_s(\cC_k,\mathcal{P}_{k}) + F_t(\cC_k,\mathcal{P}_{k})),
\end{align}
\vspace{-8pt}
\begin{align}
 \nonumber  F_s(\cC_k,\mathcal{P}_{k}) &= 
\frac{1}{|\mathcal{P}_{k}|} \sum_{\hat{\cC}_k \in \mathcal{P}_{k}} 
\cos (\cC_k,\hat{\cC}_k),
\\
 \nonumber F_t(\cC_{k},\mathcal{P}_{k}) &= 
\sigma(\max_{\hat{\cC}_k \in \mathcal{P}_{k}} 
\cos (\cC_k,\hat{\cC}_k)).
\end{align}

\vspace{2pt}
\paragraph{DP.} Solving the optimization problem and determining target scene number
can be effectively conducted by dynamic programming (DP).  
The update of $F(n=j)$ is
\[
\max_k \{F^{\star}(n=j-1 | \mC_{1:k})+g(\phi_j=\{\cC_{k+1},\dots,\cC_{|\mC|}\}) \},
\]
where $\mC_{1:k}$ is the set containing the first $k$ super shots.
\paragraph{Iterative optimization.}
The above DP could give us a scene cut result, but we can further take this result as a new super shot set and iteratively merge them to improve the final result. When the super shot updates, we also need to update these super shot representations. A simple summation over all the contained shots may not be an ideal representation for a super shot, as there are some shots containing less informations. Therefore, it would be better if we refine the representation of super shots in the optimal grouping. The details of this refinement on super shot representation are given in the supplements.


\section{Experiments}
\label{sec:exp}
\subsection{Experimental Setup}
\noindent\textbf{Data.} 
We implement all the baseline methods with our \emph{MovieScenes} dataset.
The whole annotation set is split into \textit{Train}, \textit{Val}, and \textit{Test} sets with the ratio
10:2:3 on video level. 

\vspace{6pt}
\noindent\textbf{Implementation details.} 
We take cross entropy loss for the binary classification. Since there exists unbalance in the dataset, \ie~non-scene-transition shot boundaries dominate in amount (approximate 9:1),
we take a 1:9 weight on cross entropy loss for non-scene-transition shot boundary and scene-transition shot boundary respectively.
We train these models for 30 epochs with Adam optimizer. The initial learning rate is 0.01 and the learning rate will be divided by 10 at the 15th epoch.

In the global optimal grouping, we take $j=600$ super shots from local segmentation according to the obtained classification scores for these shot boundaries (a movie usually contains $1k \sim 2k$ shot boundaries.) The range of target scenes are from $50$ to $400$, i.e. $i \in [50,400]$. These values are estimated based on the \textit{MovieScenes} statistics.

\vspace{6pt}
\noindent\textbf{Evaluation Metrics.} 
We take three commonly used metrics: 1) Average Precision (AP). Specifically in our experiment, it is the mean of AP of $o_i=1$ for each movie. 2) M$iou$: a weighted sum of intersection of union of a detected scene boundary with respect to its distance to the closest ground-truth scene boundary. 3) Recall@3s: recall at 3 seconds, the percentage of annotated scene boundaries which lies within 3s of the predicted boundary.

\begin{table*}[!t]
	\caption{Scene segmentation result. In our pipeline, Multi-Semantics means multiple semantic elements, BNet means shot boundary modeling boundary net, Local Seq means local sequence model, Global means global optimal grouping.}
	\vspace{-2pt}
	\begin{center}
		{
			\begin{tabular}{l|ccccc}
				\toprule
				Method            
				& AP ($\uparrow$) & $M_{iou}$ ($\uparrow$) & Recall($\uparrow$)& Recall@3s ($\uparrow$) \\
				\midrule
				Random guess 			 
				& 8.2 & 26.8 & 49.8  & 54.2 \\
				\midrule
				Rasheed \etal, GraphCut~\cite{rasheed2005detection} 
				& 14.1 & 29.7 & 53.7  & 57.2 \\
				Chasanis \etal, SCSA~\cite{chasanis2008scene}
				& 14.7 & 30.5 & 54.9  & 58.0 \\
				Han \etal, DP~\cite{han2011video}  
				& 15.5 & 32.0 & 55.6  & 58.4 \\
				Rotman \etal, Grouping~\cite{rotman2017optimal} 
				& 17.6 & 33.1 & 56.6  & 58.7 \\
				Tapaswi \etal, StoryGraph~\cite{tapaswi2014storygraphs} 
				& 25.1 & 35.7 & 58.4  & 59.7 \\
				Baraldi \etal, Siamese~\cite{baraldi2015deep}
				& \textbf{28.1} & \textbf{36.0} & \textbf{60.1}  & \textbf{61.2} \\
				\midrule
				LGSS (Base)  
				& 19.5 & 34.0 & 57.1  & 58.9  \\
				LGSS (Multi-Semantics) 
				& 24.3 & 34.8 & 57.6  & 59.4    \\
				LGSS (Multi-Semantics+BNet) 
				& 42.2 & 44.7 & 67.5  & 78.1      \\
				LGSS (Multi-Semantics+BNet+Local Seq) 
				& 44.9 & 46.5 & 71.4  & 77.5 \\
				LGSS (all, Multi-Semantics+BNet+Local Seq+Global) 
				& \textbf{47.1} & \textbf{48.8} &\textbf{73.6}  & \textbf{79.8} \\			
				\midrule
				Human upper-bound      
				& 81.0 & 91.0 & 94.1  & 99.5\\
				\bottomrule
			\end{tabular}
		}
	\end{center}
\vspace{-10pt}
	\label{tab:tran}
\end{table*}

\subsection{Quantitative Results}
The overall results are shown in Table~\ref{tab:tran}. 
We reproduce existing methods~\cite{rasheed2005detection,chasanis2008scene,han2011video,rotman2017optimal,tapaswi2014storygraphs,baraldi2015deep} with deep place features for fair comparison.
The base model applies temporal convolution on shots with the place feature, and we gradually add the following four modules to it, \ie, 1) multiple semantic elements (Multi-Semantics), 2) shot boundary representation at clip level (BNet), 3) coarse prediction at segment level with a local sequence model (Local Seq), and 4) global optimal grouping at movie level (Global).

\vspace{6pt}
\noindent\textbf{Analysis of overall results.}
The performance of random method depends on the ratio of scene-transition/non-scene-transition shot boundary in the test set, which is approximately $1:9$. All the conventional methods~\cite{rasheed2005detection,chasanis2008scene,han2011video,rotman2017optimal} outperform random guess, yet do not achieve good performance since they only consider the local contextual information and fail to capture semantic information. \cite{tapaswi2014storygraphs,baraldi2015deep} achieve better results than conventional methods~\cite{rasheed2005detection,chasanis2008scene,han2011video,rotman2017optimal} by considering a large range information.

\vspace{6pt}
\noindent\textbf{Analysis of our framework.}
Our base model applies temporal convolution on shots with the place feature and achieves $19.5$ on AP.
With the help of multiple semantic elements, our method improves from $19.5$ (Base) to $24.3$ (Multi-Semantics) ($24.6\%$ relatively).
The framework with shot boundary modeling using BNet raises the performance from $24.3$ (Multi-Semantics) to $42.2$ (Multi-Semantics+BNet) ($73.7\%$ relatively) which suggests that in the scene segmentation task, modeling shot boundary directly is useful.
The method with local sequence model (Multi-Semantics+BNet+Local Seq) achieves $2.7$ absolute and $6.4\%$ relative improvement than model (Multi-Semantics+BNet) from $42.2$ to $44.9$. 
The full model includes both local sequence model and global optimal grouping (Multi-Semantics+BNet+Local Seq+Global) further improves the results from $44.9$ to $47.1$, which shows that a movie level optimization are important to scene segmentation.

In all, with the help of multiple semantic elements, clip level shot modeling, segment level local sequence model, and movie level global optimal grouping, our best model outperforms base model and former best model~\cite{baraldi2015deep} by a large margin, which improves $27.6$ absolutely and $142\%$ relatively on base model (Base), and improves $19.0$ absolutely and $68\%$ relatively on Siamese~\cite{baraldi2015deep}. These verify the effectiveness of this local-to-global framework.

\begin{table}[!t]
	\caption{Multiple semantic elements scene segmentation ablation results, where four elements are studied including place, cast, action and audio.}
	\begin{center}
			\begin{tabular}{l|cccc|c}
				\toprule
				Method           &place &cast &act &aud           
				& AP ($\uparrow$)  \\
				\midrule
				Grouping~\cite{rotman2017optimal} &\checkmark&&& 
				& 17.6  \\
				StoryGraph~\cite{tapaswi2014storygraphs}&\checkmark&&&   
				& 25.1  \\
				Siamese~\cite{baraldi2015deep}  &\checkmark&&&     
				& \textbf{28.1}  \\ \midrule
				Grouping~\cite{rotman2017optimal} &\checkmark&\checkmark&\checkmark&\checkmark 
				& 23.8  \\
				StoryGraph~\cite{tapaswi2014storygraphs} &\checkmark&\checkmark&\checkmark&\checkmark   
				& 33.2  \\
				Siamese~\cite{baraldi2015deep}&\checkmark&\checkmark&\checkmark&\checkmark  
				& \textbf{34.1} \\
				\midrule \midrule
				LGSS   &&&&\checkmark                            
				& 17.5            \\
				LGSS  &&&\checkmark&                            
				& 32.1            \\
				LGSS   &&\checkmark&&                            
				& 15.9                \\
				LGSS   &\checkmark&&&                            
				& 39.0                \\ \midrule
				LGSS  &\checkmark&&&\checkmark                       
				& 43.4                \\
				LGSS  &\checkmark&&\checkmark&                      
				& 45.5                \\
				LGSS  &\checkmark&\checkmark&&                       
				& 43.0                \\ \midrule
				LGSS &\checkmark&\checkmark&\checkmark&\checkmark 
				& \textbf{47.1} 	\\	
				\bottomrule
			\end{tabular}
	\end{center}
	\label{tab:multi}
\vspace{-15pt}
\end{table}

\subsection{Ablation Studies}
\paragraph{Multiple semantic elements.}
We take the pipeline with shot boundary modeling BNet, local sequence model and global optimal grouping as the base model. As shown in Table \ref{tab:multi}, gradually adding mid-level semantic elements  improves the final results. Starting from the model using place only, audio improves $4.4$, action improves $6.5$, casts improves $4.0$, and improves $8.1$ with all together. This result indicates that place, cast, action and audio are all useful information to help scene segmentation. 

Additionally, with the help of our multi-semantic elements, other methods~\cite{rotman2017optimal,tapaswi2014storygraphs,baraldi2015deep} achieve $20\% \sim 30\%$ relative improvements. This result further justifies our assumption that multi-semantic elements contributing to the scene segmentation.

\vspace{6pt}
\noindent\textbf{Influence of temporal length.}
We choose different window sizes in the shot boundary modeling at clip level (BNet) and different sequence lengths of Bi-LSTM at segment level (Local Seq). The result is shown in Table~\ref{tab:temporal}.
The experiments show that a longer range of information improves the performance. Interestingly, the best results come from 4 shots for a shot boundary modeling and 10 shot boundaries as the input of a local sequence model, which involves 14 shot information in total. This is approximately the length of a scene. It shows that this range of temporal information is helpful to scene segmentation.

\begin{table}[]
	\caption{Comparison of different temporal window size at clip and segment level. The vertical line differs on the window size of clip level shot boundary modeling (BNet), the horizontal line differs on the length of segment level sequence model (seq.).}
	\begin{center}
	\begin{tabular}{l|cccccc}
		\toprule
		{BNet}$\backslash${seq}  & 1     & 2     & 5         & 10    & 20    \\ \midrule
		2 & 43.4 & 44.2 & 45.4 & 46.3 & \textbf{46.5} \\ \hline
		4 & 44.9 & 45.2 & 45.7 & \textbf{47.1} & 46.9  \\ \hline
		6 & 44.7 & 45.0 & 45.8 & \textbf{46.7} & 46.6  \\ \bottomrule
	\end{tabular}
	\end{center}
	\vspace{-10pt}
	\label{tab:temporal}
\end{table}

\begin{table}[]
	\caption{Comparison of different hyper-parameters in global optimal grouping and different choices of initial super shot number.}
	\begin{center}
		\begin{tabular}{l|cccc}
			\toprule
			{Iter \#}$\backslash${Init \#}& 400   & 600   & 800                       & 1000  \\ \midrule
			2 & \textbf{46.5} & 46.3  & 45.9 & 45.1 \\
			4 & 46.5 		  & 46.9  & 46.4 & 45.9 \\
			5 & 46.5 & \textbf{47.1} &  \textbf{46.6} & \textbf{46.0} \\ \midrule
			Converged value & 46.5 & {47.1} & {46.6} & 46.0 \\
			\bottomrule
		\end{tabular}
	\end{center}
	\vspace{-5pt}
	\label{tab:global}
\end{table}

\vspace{6pt}
\noindent\textbf{Choice of hyper-parameters in global optimal grouping.}
We differ the iteration number of optimization (Iter \#) and the initial super shots number (Init \#) and show the results in Table~\ref{tab:global}. 

We first take a look at each row and change the initial super shots number.
The setting with initial number $600$ achieves the best results, since it is close to the target scene number $50\sim400$ and meanwhile ensures enough large search space. 
Then, when we look at each column, we observe that the setting with initial number $400$ converges in the fastest way. It achieves the best results very quickly after $2$ iterations.
And all the settings coverge within $5$ iterations.

\subsection{Qualitative Results}
\begin{figure}[!t]
	\begin{center}
		\includegraphics[width=\linewidth]{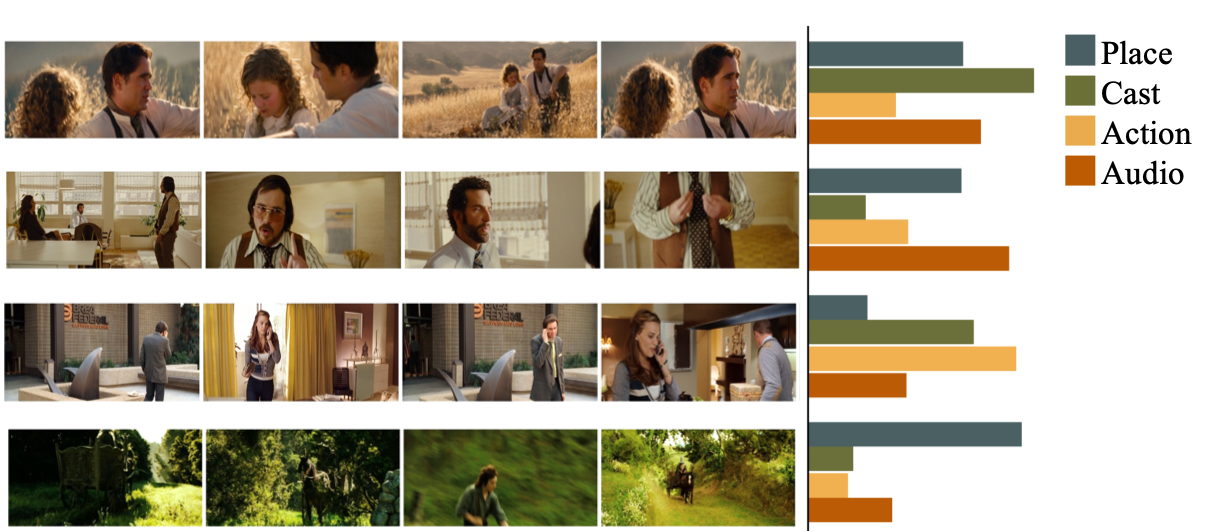}
	\end{center}
	\vspace{-5pt}
	\caption{\small
		Multiple semantic elements interpretation, where the norm of similarity of each semantic element is represented by the corresponding bar length. These four movie clips illustrate how different elements contribute to the prediction of a scene.
	}
	\label{fig:multi}
		\vspace{-0pt}
\end{figure}

Qualitative results showing the effectiveness of our multi-modal approach is illustrated in Figure~\ref{fig:multi}, and the qualitative results of global optimal grouping are shown in Figure~\ref{fig:global_vis}.~\footnote{More results are shown in the supplementary materials.}

\vspace{6pt}
\noindent\textbf{Multiple semantic elements.}
To quantify the importance of multiple semantic elements, we take the norm of the cosine similarity for each modality.
Figure~\ref{fig:multi} (a) shows an example where the cast is very similar in consecutive shots and help to contribute to the formation of a scene.
In Figure~\ref{fig:multi} (b), the characters and their actions are hard to recognize: the first shot is a long shot where the character is very small, and the last shot only shows one part of the character without a clear face. In these cases, a scene is recognized thanks to the similar audio feature that is shared among these shots. 
Figure~\ref{fig:multi} (c) is a typical ``phone call'' scene where the action in each shot is similar. 
In Figure~\ref{fig:multi} (d), only place is similar and we still conclude it as one scene. From the above observations and analysis on more such cases, we come to the following empirical conclusions: multi-modal information is complementary to each other and help the scene segmentation.

\begin{figure}[!t]
	\begin{center}
	\includegraphics[width=\linewidth]{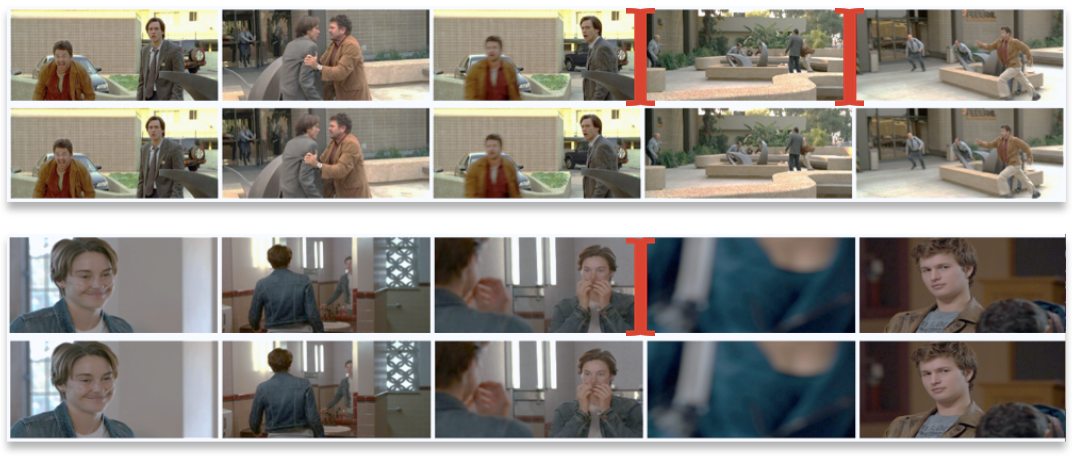}
	\end{center}
	\vspace{-10pt}
	\caption{\small
		Qualitative results of global optimal grouping in two cases. In each case, the first and second row are the results before and after the global optimal grouping respectively. The red line among two shots means there is a scene cut. The ground truth of each case is that these shots belong to the same scene. 
	}
	\label{fig:global_vis}
	\vspace{-0pt}
\end{figure}

\vspace{6pt}
\noindent\textbf{Optimal grouping.}
We show two cases to demonstrate the effectiveness of optimal grouping. There are two scenes in Figure~\ref{fig:global_vis}. 
Without global optimal grouping, a scene with sudden view point change is likely to predict a scene transition (red line in the figure), \eg~in the first case, the coarse prediction gets two scene cuts when the shot type changes from a full shot to a close shot. In the second case, the coarse prediction gets a scene cut when a extreme close up shot appears. Our global optimal grouping is able to smooth out these redundant scene cuts as we expected.

\begin{table}[!t]
	\caption{
		Scene segmentation cross dataset transfer result (AP) on existing datasets.}
	\vspace{-2pt}
	\begin{center}
		\begin{tabular}{l|cccc}
			\toprule
			Method   	 & OVSD~\cite{rotman2017optimal} & BBC~\cite{baraldi2015deep}  \\ 
			\midrule
			DP~\cite{han2011video}        			& 58.3     & 55.1 \\
			Siamese~\cite{baraldi2015deep}   		& 65.6     & 62.3 \\
			LGSS 			 & \textbf{76.2}     & \textbf{79.5} \\
			\midrule		
			DP-pretrained~\cite{han2011video}        & 62.9     & 58.7 \\
			Siamese-pretrained~\cite{baraldi2015deep}& 76.8     & 71.4 \\
			LGSS-pretrained & \textbf{85.7}     & \textbf{90.2} \\
			\bottomrule
		\end{tabular}
	\end{center}
	\vspace{-5pt}
	\label{tab:existdata}
\end{table}

\subsection{Cross Dataset Transfer}
We test different methods DP~\cite{han2011video} and Siamese~\cite{baraldi2015deep} on existing
datasets OVSD~\cite{baraldi2015deep} and BBC~\cite{rotman2017optimal} with pretraining on our MovieScenes dataset, and the results are shown in Table~\ref{tab:existdata}. With pretraining on our dataset, the performances achieve significant improvements, \ie~$\sim10$ absolute and $\sim15\%$ relative improvements in AP. 
The reason is that our dataset covers much more scenes and brings a better generalization ability to the model pretrained on it.

\section{Conclusion}
\label{sec:conclusion}
In this work, we collect a large-scale annotation set for scene segmentation on $150$ movies containing $270K$ annotations. We propose a local-to-global scene segmentation framework to cover a hierarchical temporal and semantic information.
Experiments show that this framework is very effective and achieves much better performance than existing methods.
A successful scene segmentation is able to support a bunch of movie understanding applications.~\footnote{More details are shown in the supplementary materials.}
All the studies in this paper together show that scene analysis is a challenging but meaningful topic which deserves further research efforts.

\vspace{6pt}
\noindent\textbf{Acknowledgment}
This work is partially supported by the General Research Fund (GRF) of Hong Kong (No. 14203518 \& No.
14205719) and SenseTime Collaborative Grant on Large-scale Multi-modality Analysis.

{\small
	\bibliographystyle{ieee_fullname}
	\bibliography{main}
}

\end{document}